\pdfoutput=1
\documentclass[10pt,twocolumn,letterpaper]{article}

\usepackage[pagenumbers]{cvpr} 

\definecolor{cvprblue}{rgb}{0.21,0.49,0.74}
\usepackage[pagebackref,breaklinks,colorlinks,allcolors=cvprblue]{hyperref}
\usepackage{multirow}
\usepackage{bbding}
\usepackage{pifont}
\usepackage{colortbl}
\usepackage{float}     
\usepackage{makecell}  
\usepackage[misc]{ifsym}

\def\paperID{***} 
\def\confName{CVPR}
\def\confYear{2026}

\title{AnyMS: Bottom-up Attention Decoupling for Layout-guided and Training-free Multi-subject Customization}

\author{
    Binhe Yu$^{1,*}$, Zhen Wang$^{2,*}$, Kexin Li$^{3}$, Yuqian Yuan$^{1}$, WenQiao Zhang$^{1}$, Long Chen$^{2}$, Juncheng Li$^{1}$\\
    Jun Xiao$^{1}$, Yueting Zhuang$^{1}$\\[0.5em]
    $^{1}$Zhejiang University, $^{2}$HKUST, $^{3}$Zhejiang Tobacco Monopoly Administration\\
}

\begin{document}

\twocolumn[{%
\renewcommand\twocolumn[1][]{#1}%
\maketitle
\begin{center}
    \centering
    \captionsetup{type=figure}
    \includegraphics[width=1\textwidth]{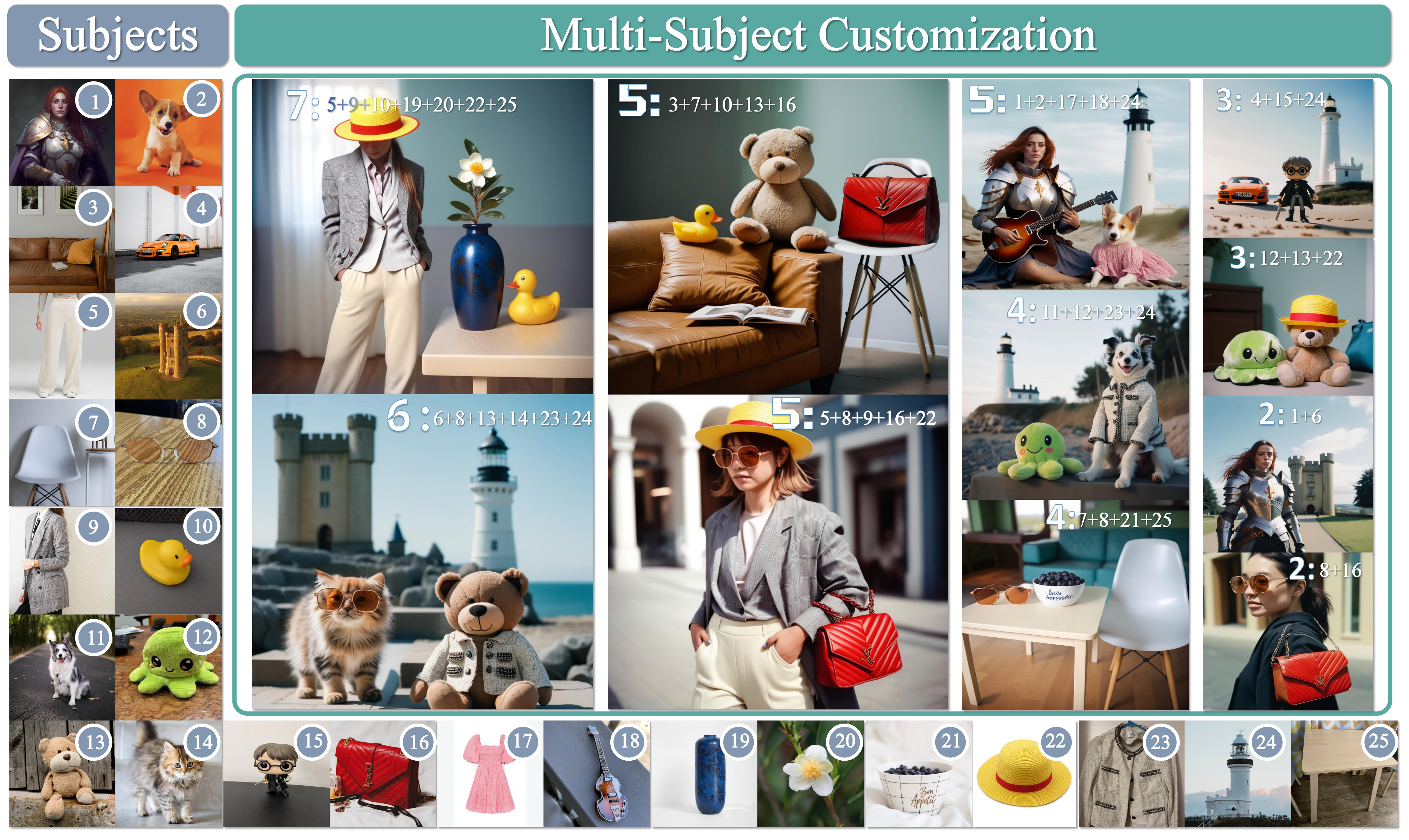}
    \vspace{-1em}
    \captionof{figure}{\small AnyMS enables \textbf{training-free} layout-guided multi-subject customization, supporting diverse subject combinations and scaling to larger numbers while maintaining a balance among layout control, text alignment, and identity preservation. See more visualization details and layout configurations in the Appendix.}
\end{center}%
}]

\renewcommand{\thefootnote}{\fnsymbol{footnote}} 
\footnotetext[1]{Equal Contribution.}           

\begin{abstract}
Multi-subject customization aims to synthesize multiple user-specified subjects into a coherent image. To address issues such as subjects missing or conflicts, recent works incorporate layout guidance to provide explicit spatial constraints. However, existing methods still struggle to balance three critical objectives: text alignment, subject identity preservation, and layout control, while the reliance on additional training further limits their scalability and efficiency. In this paper, we present \textbf{AnyMS}, a novel training-free framework for layout-guided multi-subject customization. AnyMS leverages three input conditions: text prompt, subject images, and layout constraints, and introduces a bottom-up dual-level attention decoupling mechanism to harmonize their integration during generation. Specifically, global decoupling separates cross-attention between textual and visual conditions to ensure text alignment. Local decoupling confines each subject’s attention to its designated area, which prevents subject conflicts and thus guarantees identity preservation and layout control. Moreover, AnyMS employs pre-trained image adapters to extract subject-specific features aligned with the diffusion model, removing the need for subject learning or adapter tuning. Extensive experiments demonstrate that AnyMS achieves state-of-the-art performance, supporting complex compositions and scaling to a larger number of subjects. 
\end{abstract}
    
\section{Introduction}
\label{sec:1}

\begin{figure*}[t]
    \centering
    \includegraphics[width=1\linewidth]{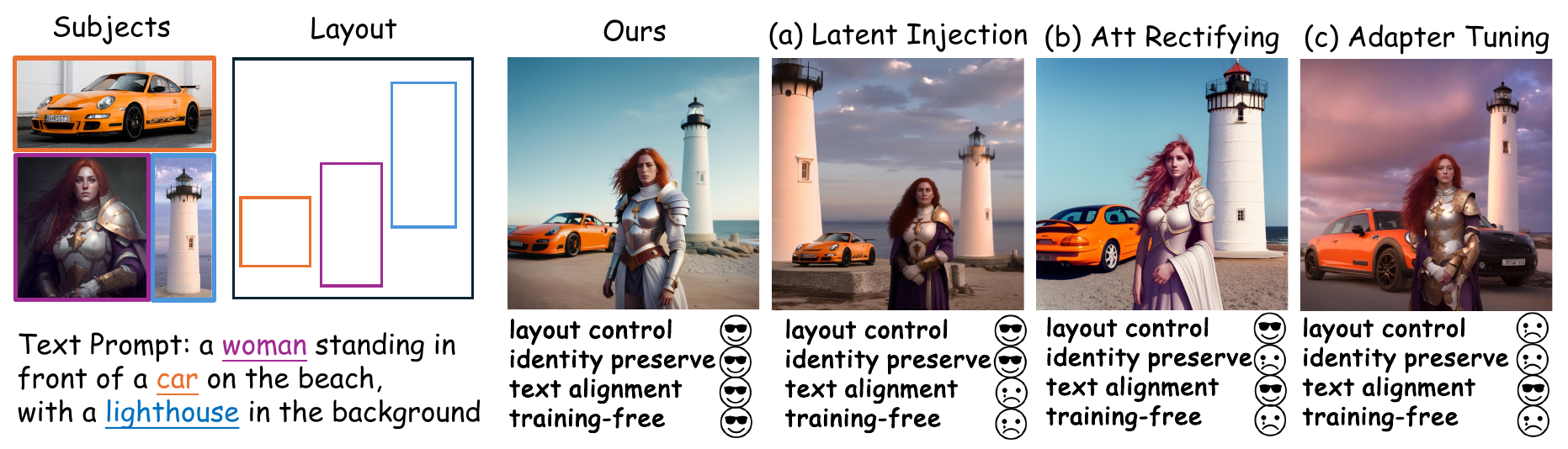}
    \vspace{-2em}
    \caption{\textbf{Layout-guided Multi-subject Customization Results.} (a) Result of latent injection method MuDI~\citep{jang2024identity}. (b) Result of attention rectifying method Cones2~\citep{liu2023customizable}. (c) Result of adapter tuning method MS-Diffusion~\citep{wang2024ms}. Different colors show the associations between subjects and their corresponding layout configurations.}
    \label{fig:2}
    \vspace{-1.5em}
\end{figure*}

Recent advances in large-scale pre-trained diffusion models~\citep{dhariwal2021diffusion,nichol2021glide,ramesh2022hierarchical,rombach2022high,saharia2022photorealistic} have enabled the novel application of customized image generation~\citep{gal2022image,ruiz2023dreambooth,chen2023disenbooth,wang2024event}, allowing users to generate images containing specific subjects of interest. While single-subject customization has achieved remarkable success~\citep{gal2022image,ruiz2023dreambooth,gal2023encoder,li2024blip}, recent research has advanced toward the more challenging task of multi-subject customization~\citep{kumari2023multi,liu2023cones,gu2024mix}. This paradigm focuses on synthesizing multiple custom subjects into coherent scenes guided by textual prompts, 
thereby offering greater flexibility and personalization in user-driven content creation.

Mainstream multi-subject customization methods have evolved through several approaches. A common line of work adopts joint training~\citep{kumari2023multi,liu2023cones} with data augmentation~\citep{han2023svdiff,jang2024identity}, optimizing the model on mixed multi-subject data. Alternatively, some methods leverage single-subject tuning with composition strategies~\citep{kong2024omg,kwon2024concept,jin2025latexblend}, where individually learned subjects (\eg, LoRAs~\citep{chen2025iteris}) are combined to form multi-subject scenes. While effective in simple settings (\eg, two subjects), these approaches struggle to scale as the number of subjects increases, often resulting in subject missing or conflict, \ie, some subjects fail to appear, or their attributes become confused. Subsequently, a new direction~\citep{liu2023customizable,gu2024mix,wang2024ms,zhu2025multibooth} has emerged to introduce additional layout guidance (\eg, bounding boxes), which aims to alleviate the above issues by explicitly constraining the spatial arrangement of subjects.


Specifically, existing layout-guided multi-subject customization methods mainly implement the layout guidance from three perspectives: \textbf{a) Latent Injection.} The typical method~\citep{jang2024identity} composes segmented subjects within bounding boxes to form an initial latent noise, injecting spatial and appearance priors. While effective for layout and identity preservation, such latent-level constraint often undermines text alignment (\eg, object relations), resulting in incoherent generation (\cf, Figure~\ref{fig:2}(a)). \textbf{b) Attention Rectifying.} These methods~\citep{liu2023customizable,chen2024training,zhu2025multibooth} adjust the cross-attention map of each subject using bounding boxes, typically by enhancing activations in target regions while suppressing irrelevant areas. However, this attention-level guidance often requires encoding each subject as a special token concatenated with the text prompt and controlled via text cross-attention, which can cause conflicts between visual and textual conditions, leading to imprecise identity preservation shown in Figure~\ref{fig:2}(b). \textbf{c) Adapter Tuning.} MS-Diffusion~\citep{wang2024ms} introduces a trainable adapter module to jointly encode visual features, text embeddings, and layout constraints. However, they rely on carefully curated layout-labeled multi-subject data with additional module tuning, which significantly increases computational cost and limits their generalization capability to unseen subjects or combinations (\cf, Figure~\ref{fig:2}(c)). In summary, existing layout-guided approaches still suffer from two major limitations:

\begin{itemize}[leftmargin=*]
\setlength{\itemsep}{0pt}

    \item \textbf{1)} Difficulty in balancing the trade-off among text alignment, subject identity preservation, and layout control, especially as the number of subjects increases.

    
    \item  \textbf{2)} Reliance on additional training for subject learning or adapter tuning, leading to strong data dependency and substantial computational overhead. 
    
\end{itemize}
    
To address these limitations, we propose \textbf{AnyMS}, a novel \emph{training-free} layout-guided multi-subject customization framework. Based on the three types of input conditions for layout-guided multi-subject customization — \emph{textual} (\eg, text prompts), \emph{visual} (\eg, subject images), and \emph{layout} (\eg, bounding boxes) — AnyMS performs a bottom-up dual-level attention decoupling to balance their integration alongside the general denoising process of diffusion generation:
\textbf{1)} \emph{Global decoupling}: separating cross-attention between text (\ie, text cross-attention) and subject images (\ie, image cross-attention) to mitigate global conflicts between textual and visual conditions, thereby ensuring text alignment.
\textbf{2)} \emph{Local decoupling}: further disentangling image cross-attention using layout constraints, where each region only attends to its corresponding subject, avoiding interference among multiple subjects, thus guaranteeing both subject identity preservation and layout control.

In addition, AnyMS employs pre-trained image adapters~\citep{ye2023ip,li2024blip} to extract subject-specific visual features aligned with the diffusion model, thereby eliminating the need for time-consuming subject learning or additional tuning. By decomposing and harmonizing textual, visual, and layout conditions, AnyMS achieves a better balance between text alignment, subject identity preservation, and layout control (\cf, Figure~\ref{fig:2}). Extensive experiments show that our method achieves state-of-the-art performance, supporting more complex and creative multi-subject customization with improved efficiency and generalizability.

In summary, we made three contributions in this paper:
{1)} We introduce \textbf{AnyMS}, a novel \emph{training-free} framework for layout-guided multi-subject customization that employs a bottom-up dual-level attention decoupling mechanism that disentangles textual, visual, and layout conditions. {2)} We incorporate pre-trained image adapters to efficiently extract subject-specific features, eliminating the need for additional tuning or subject learning.
{3)} We conduct extensive experiments across diverse benchmarks, demonstrating that AnyMS achieves state-of-the-art performance with improved efficiency, supporting complex compositions and scaling to a larger number of subjects.

\section{Related Work}
\label{sec:2}

\noindent\textbf{Text-to-image Generation.}
Text-to-image (T2I) generation aims to synthesize realistic images from textual descriptions. Early GAN-based approaches~\citep{reed2016generative,li2019controllable} have been largely surpassed by diffusion models~\citep{ho2020denoising,song2020denoising}, which progressively denoise latent variables under text guidance~\citep{dhariwal2021diffusion,ho2022classifier}. Recent advances such as Stable Diffusion~\citep{rombach2022high} and SDXL~\citep{podell2023sdxl} further improve fidelity and efficiency. While these models achieve remarkable performance on generic prompts, they cannot directly handle more complex cases of user-specific customization.

\noindent\textbf{Single-subject Customization.} Customized generation introduces user-specified subjects into T2I models with faithful identity preservation and textual description alignment. Existing single-subject customization approaches can be categorized into three groups:  
(1) \emph{Training-based}, such as DreamBooth~\citep{ruiz2023dreambooth} and Textual Inversion~\citep{gal2022image}, which fine-tune or optimize embeddings to bind subjects with special tokens or parameters;  
(2) \emph{LoRA-based}, where lightweight parameter tuning (\eg, LoRA~\citep{soboleva2025t,kong2024omg}) is employed to inject subject-specific features;  
(3) \emph{Adapter-based}, such as Blip-Diffusion~\citep{li2023blip} and SSR-Encoder~\citep{zhang2024ssr}, which learn auxiliary encoders to extract diffusion aligned subject features.  
These approaches are effective for a single subject but face scalability challenges when extended to multiple subjects.


\noindent\textbf{Multi-subject Customization.} The multi-subject setting requires generating coherent images with multiple customized subjects, where the main difficulty lies in disentangling subject features and maintaining prompt alignment. Training-based approaches, such as CustomDiffusion~\citep{kumari2023multi} and MUDI~\citep{jang2024identity}, employ joint optimization or data augmentation to improve disentanglement, but suffer from high cost and limited generalization. Inference-time methods~\citep{jiang2025mc,jin2025latexblend,kwon2024tweediemix,ding2024freecustom}, like TweedieMix~\citep{kwon2024tweediemix}, and FreeCustom~\citep{ding2024freecustom}, instead manipulate latent variables or attention maps to merge learned subjects without retraining. While reducing overhead, these methods often encounter subject omission, attribute confusion, or degraded fidelity as the number of subjects grows.


\noindent\textbf{Multi-subject Customization with Layout Control.}
To further mitigate conflicts among multiple subjects, recent studies introduce layout guidance (\eg, bounding boxes) to constrain spatial arrangement. Existing methods can be grouped into three paradigms:  
(1) \emph{Latent injection}: MuDI~\citep{jang2024identity} initializes latent codes from segmented subjects within bounding boxes, while OMG~\citep{kong2024omg} leverages layouts from non-customized images for spatial priors;  
(2) \emph{Attention rectifying}: Cones2~\citep{liu2023customizable} and Mix-of-Show~\citep{gu2023mix} enforce attention activations within target regions, helping disentangle object features;  
(3) \emph{Adapter tuning}: MS-Diffusion~\citep{wang2024ms} introduces an adapter to jointly encode subject, prompt, and layout inputs, but requires layout-labeled training data.  
Although effective in improving spatial controllability, these methods either incur heavy training costs or struggle to balance text alignment, identity preservation, and layout control. In contrast, our AnyMS achieves a better balance across the three objectives in a training-free manner, and scales naturally to complex compositions.

\section{Method}
\label{sec:3}
\begin{figure*}[t]
    \centering
    \includegraphics[width=\linewidth]{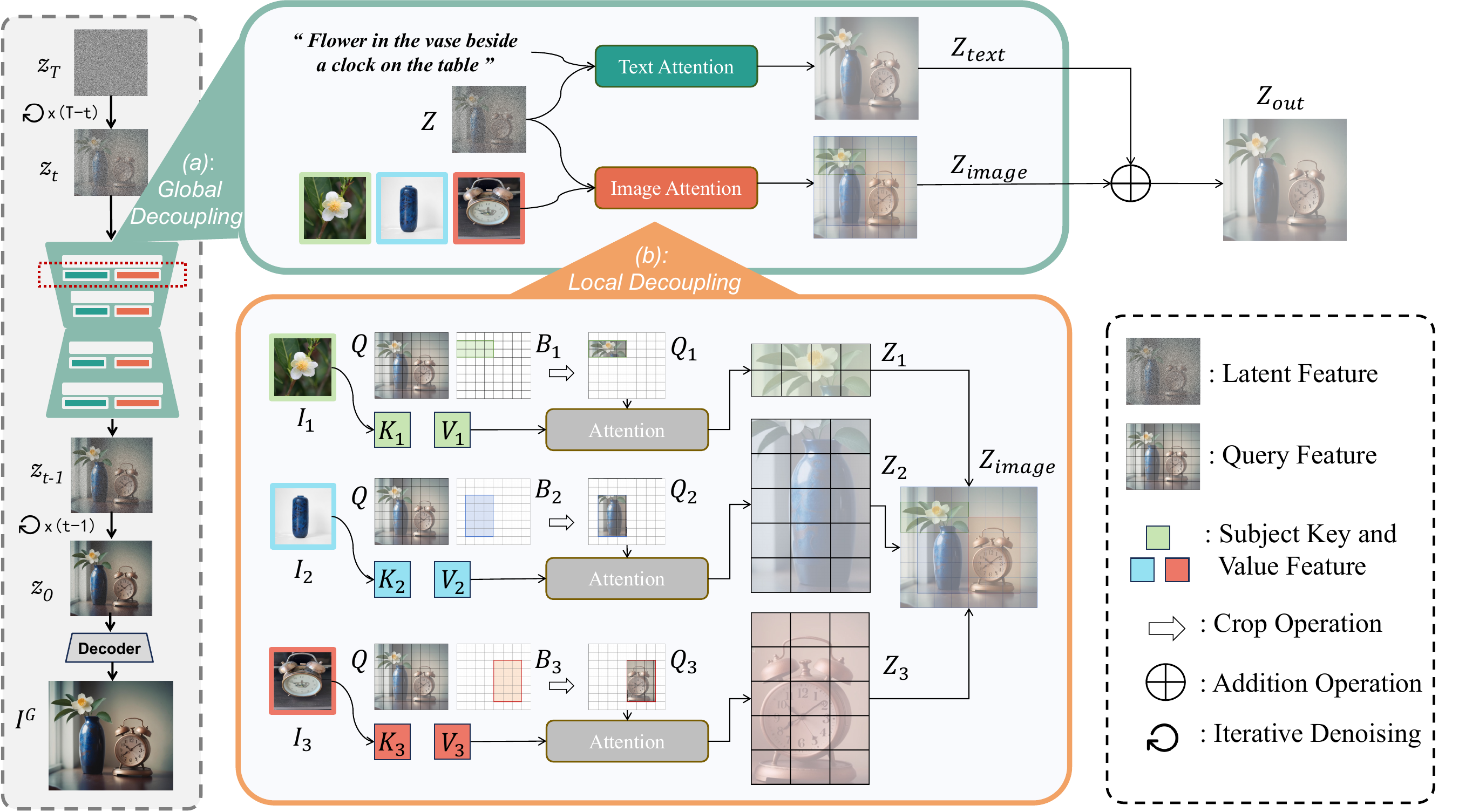}
    \vspace{-1.5em}
    \caption{\textbf{The Overview of Pipeline.} AnyMS applies a dual-level attention decoupling strategy alongside the general denoising process of the diffusion model. (a) The global decoupling separates cross-attention between text and subject images. (b) The local decoupling further disentangles image cross-attention based on layout constraints. The final $z_0$ is then decoded back to target image $I^G$.}
    \label{fig:method}.
    \vspace{-2.5em}
\end{figure*}

\subsection{Preliminary}
\noindent\textbf{Stable Diffusion Model.} Stable Diffusion Model is the text-to-image representative generative model consisting of three main components: an autoencoder $(\mathcal{E}(\,\cdot\,), \mathcal{D}(\,\cdot\,))$, a denoising network $\varepsilon_\theta(\,\cdot\,)$ and a text encoder CLIP $\tau_{\theta}(\,\cdot\,)$ ~\citep{radford2021learning}. Given an image $x$ and a text prompt $P$, the autoencoder maps the image from the pixel space to the latent space $z_0 = \mathcal{E}(x)$, while the CLIP encoder encodes the prompt to the conditional embedding $c_t = \tau_{\theta}(P)$. For the forward process, starting from $z_0$, sampled random Gaussian noise $\varepsilon \sim \mathcal{N}(0, 1)$ is applied to $z_0$ to get $z_t=\sqrt{\bar{\alpha}_t}z_0+\sqrt{1 - \bar{\alpha}_t}\epsilon$ at timestep $t$, where $\alpha$ is predefined coefficient provided by noise scheduler. For the backward process, the diffusion model is trained conditioned on the current latent $z_t$, timestep $t$ and text prompt conditions $c_t$ to predict the added noise. The model is trained with the following reconstruction loss:
\begin{equation}
    L_{\text{rec}} = \mathbb{E}_{z, \varepsilon \sim \mathcal{N}(0,1), t, c_t} \left\| \varepsilon - \varepsilon_{\theta}\left(z_t, t, c_t\right) \right\|_2^2
    \label{eq:Lrec}
\end{equation}
z\textsubscript{t} is progressively denoised to obtain z\textsubscript{0}, after which the decoder maps it back to pixel space $x = \mathcal{D}(z_0)$.

\noindent\textbf{Cross Attention Mechanism} Stable Diffusion Model utilizes U-Net~\citep{ronneberger2015u} as the backbone, which contains the cross-attention modules to integrate the text prompt. Specifically, given the latent features $Z$, the output $Z_\mathrm{out}$ of cross-attention mechanism can be formulated as:
\begin{equation}
Z_\mathrm{out} = \mathrm{{CA}_{text}}(Z, P) = \text{Softmax}\left( \frac{QK^T}{\sqrt{d}} \right) V
\label{eq:textCA}
\end{equation}
where $Q=W_qZ$ is the query attained from latent features, and $K=W_kc_t$, $V=W_vc_t$ are key and value projected from text features $c_t$, with their corresponding pretrained matrices $W_q$,$W_k$,$W_v$

\subsection{Task Definition}
We formally define the task of layout-guided multi-subject customization as follows. Given a global text prompt $P$ describing the desired scenario, and a set of subject image–layout pairs $D = \{(I_j, B_j)\}_{j=1}^{n}$, where $I_j$ is the reference image of subject $S_j$ and $B_j$ is a bounding box specifying its target position, the goal is to synthesize a compositionally coherent image $I^G$ that simultaneously achieves:  
{1)} {Textual alignment} with the text prompt $P$.  
{2)} {Subject identity preservation} for each $S_j$.  
{3)} {Layout control} by placing each subject at its designated location $B_j$.

\subsection{Approach}

\noindent\textbf{Overview.}  
We now introduce our training-free framework for layout-guided multi-subject customization. The core idea is to decouple the integration of three input conditions — \emph{textual} (text prompt), \emph{visual} (subject images), and \emph{layout} (bounding boxes) — through a bottom-up dual-level attention decoupling strategy. As illustrated in Figure~\ref{fig:method}, the generation of $I^G$ begins with a randomly initialized latent $z_T \sim \mathcal{N}(0, \mathbf{I})$, which is progressively denoised into $z_0$ by the diffusion process. During denoising, \textbf{a)} the \emph{global decoupling} separates cross-attention between text and subject images, and  
\textbf{b)} \emph{local decoupling} further disentangles image cross-attention based on layout constraints, ensuring that each spatial region attends only to its designated subject. Meanwhile, AnyMS employs a pre-trained image adapter~\citep{ye2023ip,li2024blip} to extract subject-specific features. Finally, the refined latent $z_0$ is decoded into the target image $I^G$.

\noindent\textbf{Global Decoupling.}  
A common practice in existing methods~\citep{ruiz2023dreambooth,liu2023customizable,chen2024training,zhu2025multibooth} is to introduce a special token to represent the target subject and fine-tune the model to bind the token with specific visual features. These approaches typically concatenate such learned tokens with the text prompt and process them jointly through the text cross-attention. However, this design leads to several issues. First, it inevitably causes conflicts between textual and visual conditions: the subject token is entangled with surrounding textual tokens during denoising, resulting in identity distortion, background leakage, and unintended attribute transfer. Second, when spatial relations are described in the prompt (\eg, ``\texttt{standing in front of}"), the shared cross-attention may force the subject token to compete with layout conditions, which further undermines precise position control. To mitigate these conflicts and achieve balanced condition integration, AnyMS introduces \emph{global decoupling} as shown in Figure~\ref{fig:method}(a), where text prompt and subject images are processed by separate cross-attention streams. In this way, textual semantics and visual identity are disentangled at the global level, enabling the model to preserve subjects faithfully while maintaining accurate text alignment. More detailed analysis can be found in the Appendix. The output of each cross-attention block is reformulated as:
\begin{equation}
Z_\mathrm{out} = Z_\mathrm{text} + Z_\mathrm{image} 
\end{equation}
where $Z_\mathrm{text}=\mathrm{{CA}_{text}}(Z, P)$ is obtained through the text cross-attention (\cf,Eq~\ref{eq:textCA}). And $Z_\mathrm{image}$ is obtained through the image cross-attention, which we will specify below.

\noindent\textbf{Local Decoupling.} In general image cross-attention~\citep{ding2024freecustom}, the entire latent features attend to all subject images simultaneously. This design easily leads to subject–subject conflicts, where visual features from different subjects interfere with each other, causing identity confusion or attribute mixing. To address this issue and better exploit layout constraints, as shown in Figure~\ref{fig:method}(b), AnyMS further introduces \emph{local decoupling}. Specifically, the bounding boxes are used to restrict the interaction between latent regions and their corresponding subject features, ensuring that each spatial area only attends to its designated subject. We formulate the image cross-attention as:
\begin{equation}
Z_\mathrm{image} = \mathrm{{CA}_{image}}(Z, \{(I_j, B_j)\}_{j=1}^{n})
   \label{eq:imageCA}
\end{equation}
Specifically, the local image cross-attention involves two steps.

\noindent\emph{\underline{1) Training-free Subject Feature Extraction.}}  
Instead of fine-tuning the diffusion model to learn new subject embeddings, AnyMS leverages pre-trained image adapters to directly extract subject features in a training-free manner. Given a reference image $I_j$ of subject $S_j$, the adapter encodes its image features $c_{j}$, which is already aligned with the diffusion model. This image features is then projected into the subject-specific key and value features, $K_{j} = W_k' c_{j}$ and $V_{j} = W_v' c_{j}$, using the adapter’s pretrained projection matrices $W_k'$ and $W_v'$.

\noindent\emph{\underline{2) Attention Cropping and Merging.}}
We begin by initializing $Z_\mathrm{image}$ with the global query feature $Q \in \mathbb{R}^{H\times W}$ projected from the latent features. To incorporate subject-specific information while preserving layout constraints, we adopt crop–and–merge. For each subject $S_j$ with bounding box $B_j = [h_s,h_e]\times[w_s,w_e]$, we extract the corresponding subregion $Q_{j} = Q[h_s:h_e, w_s:w_e]$. Subject features are then injected into this region via cross-attention:
\begin{equation}
Z_{j} = \text{Softmax}\left( \frac{Q_{j}K_{j}^T}{\sqrt{d}} \right)V_{j}
\label{eq:attention}
\end{equation}

Afterwards, we merge the local outputs back to their original positions:

\begin{equation}
Z_\mathrm{image}[h_s:h_e, w_s:w_e] = Z_{j}
\label{eq:merge}
\end{equation}

For overlapping regions, we resolve subject–subject conflicts by enforcing a semantic priority order (\eg, $\text{attribute} > \text{object}$, and $\text{foreground} > \text{background}$), ensuring consistent layout and identity preservation. The resulting $Z_\mathrm{image}$ thus captures local subject fidelity and structural coherence. By combining it with text cross-attention outputs $Z_\mathrm{text}$, we achieve a balanced alignment between global textual semantics and subject-aware layout.

Finally, this dual-level attention decoupling is applied to all cross-attention layers of the U-Net and every denoising timesteps during inference, yielding stronger text-subject–layout control with only marginal inference overhead. Visualized attention maps are provided in the Appendix for further insight.

\section{Experiments}
\label{sec:4}

\begin{table*}[t]
\renewcommand{\arraystretch}{1.1}
\setlength{\tabcolsep}{2pt}
    \centering
        \begin{tabular}{l|cc|ccc|c|c|c|c}
            \hline
            \multirow{2}{*}{\centering\textbf{Model}} & 
            \multicolumn{2}{c|}{\textbf{Layout Control}}  & 
            \multicolumn{3}{c|}{\textbf{Identity Preservation}}  & 
            \multicolumn{1}{c|}{\textbf{Text Alignment}} & 
            \multirow{2}{*}{\textbf{Infer Time}} &
            \multirow{2}{*}{\textbf{Memory Cost}} &
            \multirow{2}{*}{\textbf{Training}}\\
            & \textbf{AP50$\uparrow$} & \textbf{mIOU$\uparrow$}  & \textbf{CLIP-I$\uparrow$} & \textbf{DreamSim$\uparrow$} & \textbf{DINO$\uparrow$} & \textbf{CLIP-T$\uparrow$} & &\\
            \hline
            \textcolor{gray}{LatexBlend~\citep{jin2025latexblend}} & \textcolor{gray}{-} & \textcolor{gray}{-} & \textcolor{gray}{65.90} & \textcolor{gray}{43.37} & \textcolor{gray}{40.30} & \textcolor{gray}{32.40} & \textcolor{gray}{130s} & \textcolor{gray}{21GB} & \textcolor{gray}{\checkmark} \\
            Cones 2~\citep{liu2023customizable} & 26.18 & 38.10 & 67.50 & 46.00 & 41.71 & 33.49 & 108s & 26GB & \checkmark\\
            MuDI~\citep{jang2024identity} & 19.63 & 36.08 & 73.24 & 58.87 & 55.19 & 32.36 & 10s & 11GB & \checkmark \\
            MS-Diffusion~\citep{wang2024ms} & 32.26 & 48.37 & 72.04 & 59.22 & \textbf{57.33} & 34.63 & 12s & 14GB &  \ding{55}\\
            AnyMS (Ours) & \textbf{35.65} & \textbf{49.75} & \textbf{74.46} & \textbf{59.62} & 54.64 & \textbf{35.82} & 19s & 11GB &  \ding{55}\\
            \hline
        \end{tabular}
    \vspace{-0.5em}
    \caption{\textbf{Quantitative Results of Layout-guided Multi-subject Customization.} \textbf{Bold} represent the highest metric. Since LatexBlend is implemented without layout control, it is not strictly comparable and we shade its results in gray.}
    \vspace{-1.5em}
    \label{tab:performance}
\end{table*}
\subsection{EXPERIMENTAL SETUP}
\label{sec:4.1}
\noindent\textbf{Dataset.} For a fair and comprehensive evaluation, we follow the widely recognized subject customization benchmarks and conduct experiments on subjects drawn from DreamBooth~\citep{ruiz2023dreambooth}, Custom-Concept101~\citep{kumari2023multi}, and Textual Inversion~\citep{gal2022image}. In total, we collect 29 subjects covering diverse categories, including animals, objects, vehicles, and humans, which ensures broad coverage for multi-subject evaluation. To enrich diversity, we form 11 combinations for the quantitative study and present more combinations for visual display . 

\noindent\textbf{Evaluation Metrics.} We comprehensively evaluate the performance of layout-guided multi-subject customization from three perspectives:
\emph{1) Layout control.} We assess the consistency between the generated layout and the input bounding boxes by computing mean Intersection-over-Union (mIoU) and AP@50 scores. Both metrics are measured using pre-trained object detection models GroundingDINO~\citep{liu2024grounding}.
\emph{2) Identity preservation.} To evaluate how well the generated images retain subject identity, we extract the target subject regions from generated images via GroundingDINO~\citep{liu2024grounding} and SAM~\citep{kirillov2023segment}, and compute similarity with corresponding reference images using multiple image-level similarity metrics, including CLIP-I~\citep{radford2021learning}, DreamSim~\citep{fu2023dreamsim}, and DINO~\citep{oquabdinov2}.
\emph{3) Text alignment.} We evaluate the semantic consistency between generated images and text prompts using CLIP-T~\citep{radford2021learning} similarity in the CLIP embedding space. 
\emph{4) Inference overhead.} For consistency, all inference experiments are conducted on the same GPU device, and we report the average GPU memory usage and inference time required to generate one image.

\noindent\textbf{Baselines.} We compare AnyMS with representative state-of-the-art layout-guided multi-subject customization methods, which can be broadly divided into three categories:
\emph{1) Latent Injection.} \textbf{MuDI}\citep{jang2024identity} augments training data using OWLv2\citep{minderer2023scaling} and SAM~\citep{kirillov2023segment}, and introduces a latent initialization strategy with bounding boxes to provide a better starting point for inference.
\emph{2) Attention Rectifying.} \textbf{Cones2}\citep{liu2023customizable} learns subject tokens via finetuning and constrains their cross-attention maps within the assigned bounding box regions.
\emph{3) Adapter Tuning.} \textbf{MS-Diffusion}\citep{wang2024ms} trains a layout-aware adapter to jointly encode subject, prompt, and layout inputs, leveraging a carefully curated multi-subject dataset with bounding box annotations. In addition, we also report results of the state-of-the-art layout-free multi-subject customization method, \textbf{LatexBlend}~\citep{jin2025latexblend}, for reference.

\noindent\textbf{Implementation Details.} We implement our method with Stable Diffusion XL (SDXL)~\citep{podell2023sdxl} as the base model and employ IP-Adapter~\citep{ye2023ip} as the pretrained image adapter for subject feature extraction. For all baseline methods, we follow their default settings. All tested images are generated with a resolution of 1024x1024.


\label{sec:4.1}
\begin{figure}
    \centering
    \includegraphics[width=1\linewidth]{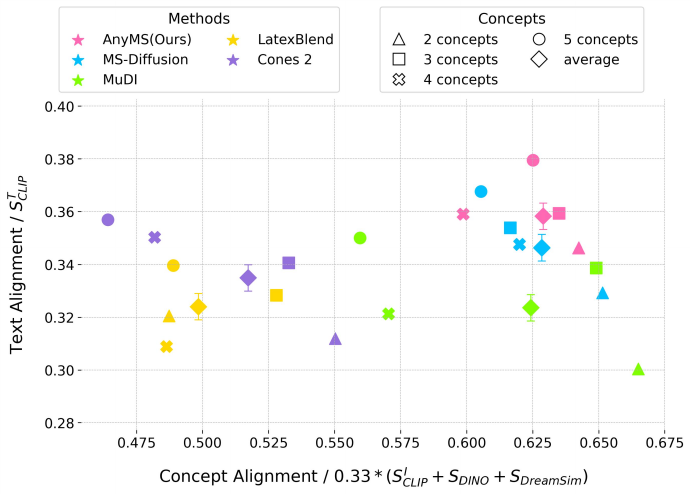}
    \vspace{-2em}
    \caption{\textbf{Detailed Quantitative Results with Different Numbers of Subjects.} Marker shape indicates the number of subjects, and color represents the method used. Rhombus denotes aggregated results.}
    \label{fig:detail}
    \vspace{-2em}
\end{figure}

\label{sec:4.1}
\begin{figure*}[t]
    \centering
    \includegraphics[width=1\linewidth]{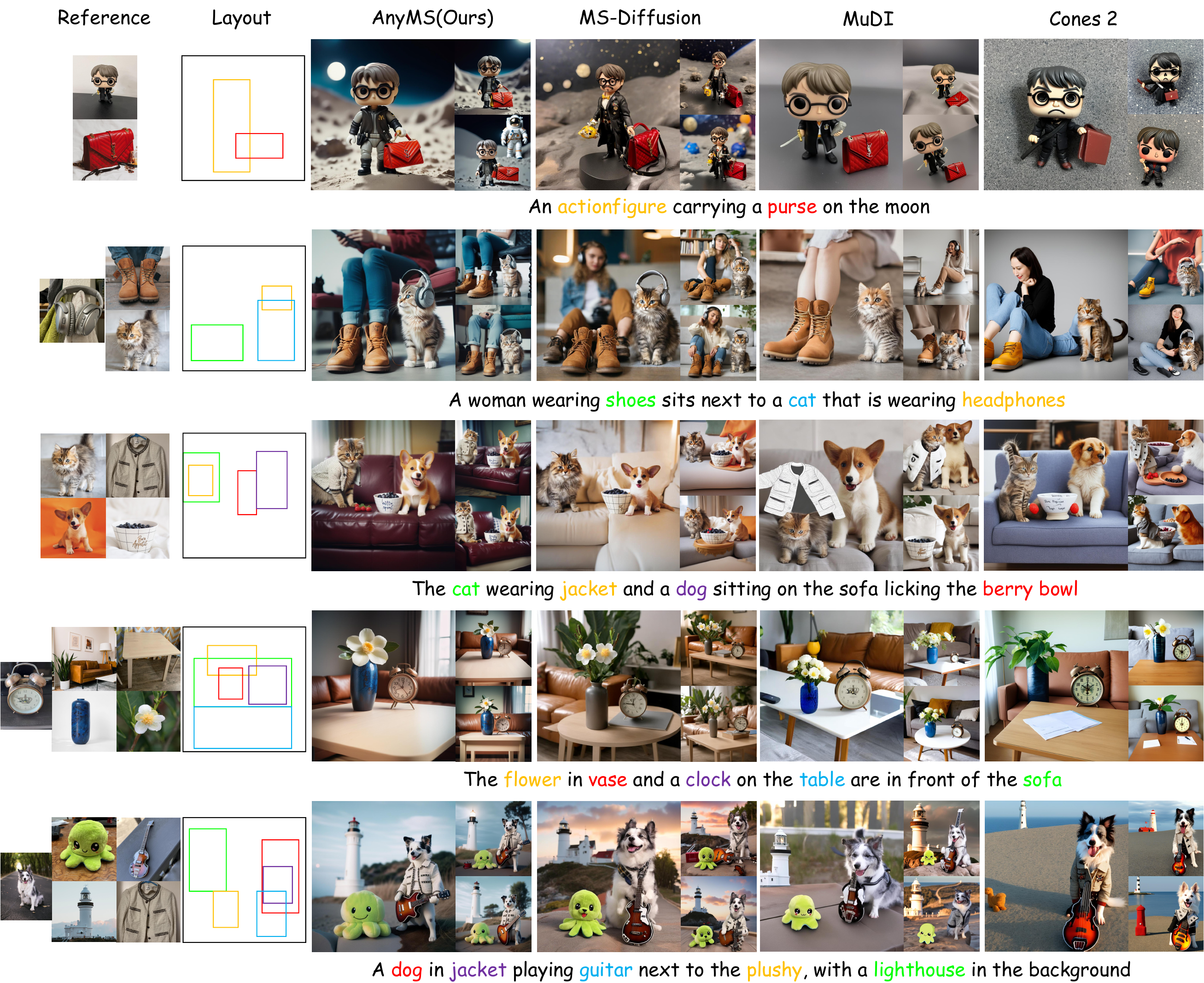}
    \vspace{-1.5em}
    \caption{\textbf{Comparison of Layout-guided Multi-subject Customization.} Different colors show the associations between subjects and their corresponding layout configurations.}
    \label{fig:qualitative}
    \vspace{-1.5em}
\end{figure*}

\subsection{Quantitative evaluation}
\label{sec:4.1}
\textbf{Setting.} We construct 11 experimental cases of multi-subject composition, covering subject counts ranging from 2 to 5, each with diverse layout configurations (\ie, bounding boxes). For every case, we randomly generate 100 images and report the averaged evaluation metrics in Table~\ref{tab:performance}. To further analyze scalability, Figure~\ref{fig:detail} visualizes the performance trends of different methods under varying numbers of subjects. See more details in Appendix.

\noindent\textbf{Results.} As shown in Table~\ref{tab:performance}, we observe: 1) For layout control and text alignment, AnyMS achieves the highest performance across both metrics, demonstrating accurate spatial control of subjects and faithful adherence to textual descriptions. 2) AnyMS also consistently surpasses baseline methods in identity preservation, achieving superior similarity scores on both CLIP-I and DreamSim, and competitive performance on the DINO score. These results demonstrate that AnyMS achieves a well-balanced trade-off among layout control, text alignment, and identity preservation. 3) Additionally, as shown in Figure~\ref{fig:detail}, AnyMS maintains strong performance when the number of subjects increases from 2 to 5. In particular, our method shows clear advantages under more complex compositions. 4) In terms of Inference overhead, AnyMS achieves superior visual quality with minimal GPU memory consumption and notably fast generation speed. These results highlight that AnyMS not only supports complex scene composition but also scales effectively to larger numbers of subjects, achieving the best overall performance. 


\subsection{Qualitative evaluation}

As shown in Figure~\ref{fig:qualitative}, for multi-subject customization cases with subject numbers varying from 2 to 5, common failure patterns across baselines include object omission, fidelity degradation, and poor prompt alignment. Specifically, MS-Diffusion struggles with complex scenarios and generalizes poorly to unseen subjects, leading to degraded identity reconstruction and inadequate prompt adherence. MuDI prioritizes identity preservation at the expense of prompt compliance and fails to model interactions between objects (\eg, \texttt{carrying} and \texttt{wearing}). Cones2 can generate multiple subjects but suffers from low identity preservation. In contrast, AnyMS excels in multi-subject customization, producing visually harmonious images with coherent layout by achieving both high fidelity and prompt alignment, while maintaining a balanced trade-off between identity preservation and layout control. Moreover, AnyMS demonstrates robust performance as the number of subjects increases, further highlighting its scalability in complex compositions.



\begin{figure*}[t]
    \centering
    \includegraphics[width=\linewidth]{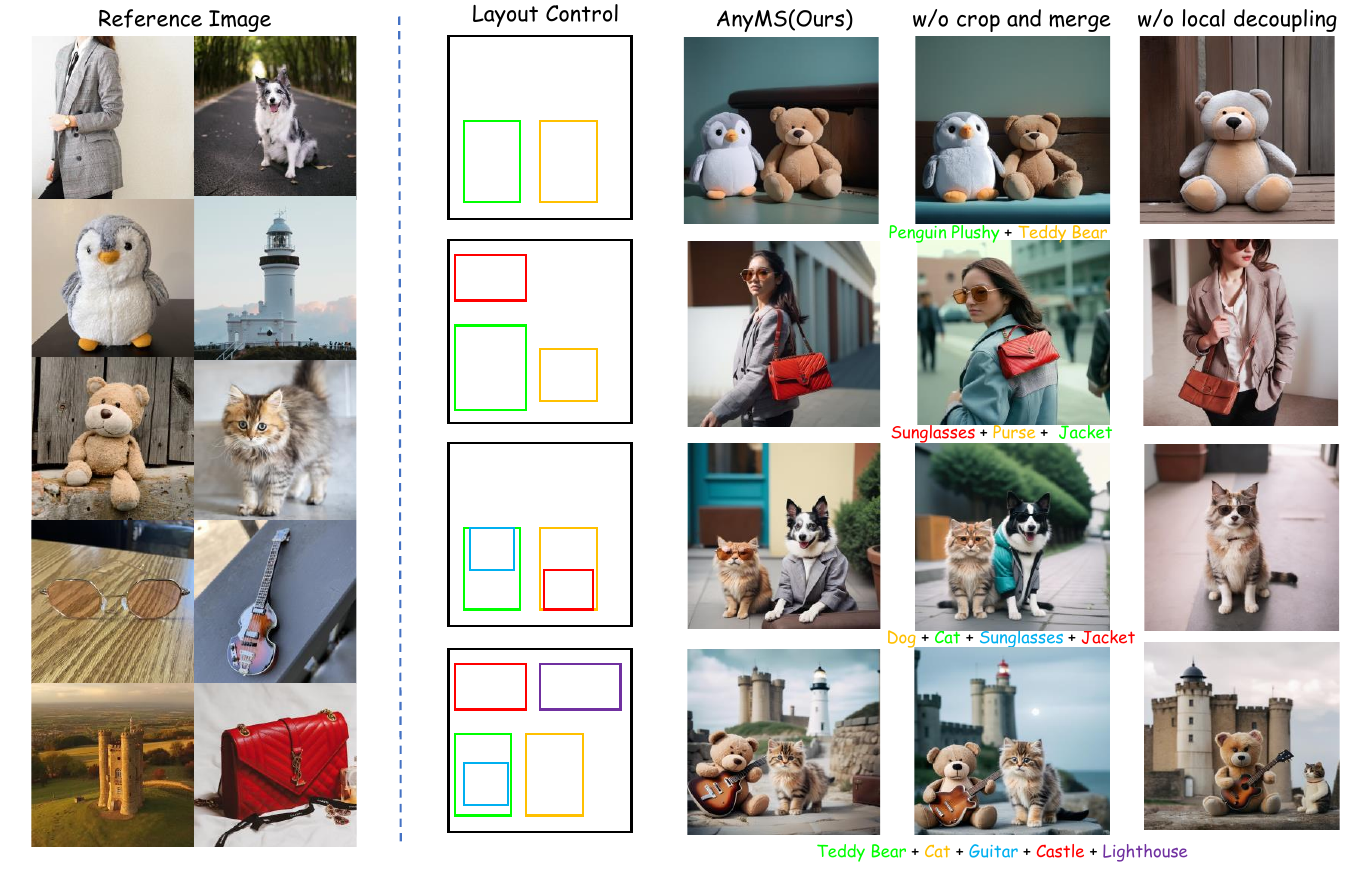}
    \vspace{-2.5em}
    \caption{\textbf{Ablations Results.} Different colors show the associations between subjects and their corresponding layout configurations. Crop-and-merge plays a crucial role in decoupling features and layout control as the number of subjects increases.}
    \label{fig:ablation}
    \vspace{-1em}
\end{figure*}

\subsection{Ablation Study}
\label{sec:4.3}

\noindent\textbf{Settings.} We conduct ablations to validate the effects of the proposed method with two different settings. 1) Remove the crop-and-merge operation in local decoupling. We directly utilize the entire query $Q$ to calculate the image attention $Z_j$ for subject $S_j$, rather than extracting the subregion $Q_j$. We then apply a mask $M_j$ based on the bounding box $B_j$ to $Z_j$ to get layout-aware attention, and add them together to get the final output $Z_\mathrm{image}$. The Eq~\ref{eq:attention} and Eq~\ref{eq:merge} are reformed as:
\begin{equation}
Z_{j} = \text{Softmax}\left( \frac{QK_{j}^T}{\sqrt{d}} \right)V_{j},  Z_\mathrm{image} = \sum_{j=1}^{n}(Z_j\odot M_j)
\label{eq:ab}
\end{equation}
2) Remove the whole local decoupling operation. We further remove all layout guidance by directly adding each $S_j$ to get the final output $Z_\mathrm{image} = \sum_{j=1}^{n}Z_j$.
\begin{table}
    \centering
    \small
        \begin{tabular}{l|c|c|c}
            \hline
             \textbf{Model} & \textbf{mIOU$\uparrow$}  & \textbf{CLIP-I$\uparrow$} &  \textbf{CLIP-T$\uparrow$} \\
            \hline
            w/o local decoupling & - & 67.26 & 35.38   \\
            w/o crop and merge & 38.78 & 72.13 & 35.67   \\
            AnyMS (Ours) & \textbf{44.55} & \textbf{73.45} & \textbf{36.50} \\
            \hline
        \end{tabular}
    \vspace{-0.5em}
    \caption{\textbf{Quantitative Results for Ablation Study.} Evaluated on seven combinations with more than three subjects.}
    \vspace{-3em}
    \label{tab:ablation performance}
\end{table}

\noindent\textbf{Results.} As shown in Figure~\ref{fig:ablation}, we have the following observations: 1) Removing the crop-and-merge operation works reasonably when the subject number is small (\eg, two subjects), but as the count increases the model struggles to disentangle subject features, leading to issues like subject missing, object fusion (\eg, the castle and the lighthouse), and degraded identity preservation (\eg, the jacket). 2) Without local decoupling, image fidelity drops further, and subject entanglement becomes more severe. In contrast, AnyMS effectively disentangles subject features and preserves high image fidelity. To further validate the effectiveness of AnyMS, we quantitatively evaluated it on scenarios involving three or more subjects with a total of seven combinations. As 
shown in Table~\ref{tab:ablation performance}, AnyMS tops all metrics. With layout guidance, the local decoupling operation further confines subjects to their designated regions, enabling a stable balance between identity preservation and layout control, while scaling robustly to larger numbers of subjects.

\section{Conclusion}
\label{sec:5}

In this paper, we introduced AnyMS, a novel training-free framework for layout-guided multi-subject customization. By performing bottom-up dual-level attention decoupling, AnyMS effectively disentangles textual, visual, and layout conditions, achieving a better balance among text alignment, subject identity preservation, and layout control. Extensive experiments and ablations validate the effectiveness of AnyMS and its scalability to increasingly complex multi-subject compositions. Moving forward, we plan to 1) extend AnyMS into video customization; 2) explore advanced techniques for customization that jointly consider subject, action, and style.

\noindent\textbf{Limitations.} While AnyMS achieves strong performance without additional training, its effectiveness still depends on the capacity of the underlying pre-trained diffusion model and image adapters. Consequently, the upper bound on the number of subjects, the complexity of scenes, and the robustness of subject feature extraction may be constrained. Future work could explore integrating stronger foundation models and adaptive feature learning strategies to further enhance scalability and generalization.

{
    \small
    \bibliographystyle{ieeenat_fullname}
    \bibliography{main}

@String(TOG= {ACM Trans. Graph.})

@String(AAAI = {AAAI})

@String(TOG   = {ACM TOG})

@article{ho2020denoising,
  title={Denoising diffusion probabilistic models},
  author={Ho, Jonathan and Jain, Ajay and Abbeel, Pieter},
  journal={Advances in neural information processing systems},
  volume={33},
  pages={6840--6851},
  year={2020}
}

@inproceedings{ruiz2023dreambooth,
  title={Dreambooth: Fine tuning text-to-image diffusion models for subject-driven generation},
  author={Ruiz, Nataniel and Li, Yuanzhen and Jampani, Varun and Pritch, Yael and Rubinstein, Michael and Aberman, Kfir},
  booktitle={Proceedings of the IEEE/CVF conference on computer vision and pattern recognition},
  pages={22500--22510},
  year={2023}
}

@inproceedings{kumari2023multi,
  title={Multi-concept customization of text-to-image diffusion},
  author={Kumari, Nupur and Zhang, Bingliang and Zhang, Richard and Shechtman, Eli and Zhu, Jun-Yan},
  booktitle={Proceedings of the IEEE/CVF conference on computer vision and pattern recognition},
  pages={1931--1941},
  year={2023}
}

@inproceedings{radford2021learning,
  title={Learning transferable visual models from natural language supervision},
  author={Radford, Alec and Kim, Jong Wook and Hallacy, Chris and Ramesh, Aditya and Goh, Gabriel and Agarwal, Sandhini and Sastry, Girish and Askell, Amanda and Mishkin, Pamela and Clark, Jack and others},
  booktitle={International conference on machine learning},
  pages={8748--8763},
  year={2021},
  organization={PmLR}
}

@inproceedings{liu2024grounding,
  title={Grounding dino: Marrying dino with grounded pre-training for open-set object detection},
  author={Liu, Shilong and Zeng, Zhaoyang and Ren, Tianhe and Li, Feng and Zhang, Hao and Yang, Jie and Jiang, Qing and Li, Chunyuan and Yang, Jianwei and Su, Hang and others},
  booktitle={European conference on computer vision},
  pages={38--55},
  year={2024},
  organization={Springer}
}

@inproceedings{kirillov2023segment,
  title={Segment anything},
  author={Kirillov, Alexander and Mintun, Eric and Ravi, Nikhila and Mao, Hanzi and Rolland, Chloe and Gustafson, Laura and Xiao, Tete and Whitehead, Spencer and Berg, Alexander C and Lo, Wan-Yen and others},
  booktitle={Proceedings of the IEEE/CVF international conference on computer vision},
  pages={4015--4026},
  year={2023}
}

@article{wang2024ms,
  title={Ms-diffusion: Multi-subject zero-shot image personalization with layout guidance},
  author={Wang, Xierui and Fu, Siming and Huang, Qihan and He, Wanggui and Jiang, Hao},
  journal={arXiv preprint arXiv:2406.07209},
  year={2024}
}

@inproceedings{jin2025latexblend,
  title={LatexBlend: Scaling multi-concept customized generation with latent textual blending},
  author={Jin, Jian and Yu, Zhenbo and Shen, Yang and Fu, Zhenyong and Yang, Jian},
  booktitle={Proceedings of the Computer Vision and Pattern Recognition Conference},
  pages={23585--23594},
  year={2025}
}

@article{jang2024identity,
  title={Identity decoupling for multi-subject personalization of text-to-image models},
  author={Jang, Sangwon and Jo, Jaehyeong and Lee, Kimin and Hwang, Sung Ju},
  journal={Advances in Neural Information Processing Systems},
  volume={37},
  pages={100895--100937},
  year={2024}
}

@article{minderer2023scaling,
  title={Scaling open-vocabulary object detection},
  author={Minderer, Matthias and Gritsenko, Alexey and Houlsby, Neil},
  journal={Advances in Neural Information Processing Systems},
  volume={36},
  pages={72983--73007},
  year={2023}
}

@article{nichol2021glide,
  title={Glide: Towards photorealistic image generation and editing with text-guided diffusion models},
  author={Nichol, Alex and Dhariwal, Prafulla and Ramesh, Aditya and Shyam, Pranav and Mishkin, Pamela and McGrew, Bob and Sutskever, Ilya and Chen, Mark},
  journal={arXiv preprint arXiv:2112.10741},
  year={2021}
}

@article{ramesh2022hierarchical,
  title={Hierarchical text-conditional image generation with clip latents},
  author={Ramesh, Aditya and Dhariwal, Prafulla and Nichol, Alex and Chu, Casey and Chen, Mark},
  journal={arXiv preprint arXiv:2204.06125},
  volume={1},
  number={2},
  pages={3},
  year={2022}
}

@article{saharia2022photorealistic,
  title={Photorealistic text-to-image diffusion models with deep language understanding},
  author={Saharia, Chitwan and Chan, William and Saxena, Saurabh and Li, Lala and Whang, Jay and Denton, Emily L and Ghasemipour, Kamyar and Gontijo Lopes, Raphael and Karagol Ayan, Burcu and Salimans, Tim and others},
  journal={Advances in neural information processing systems},
  volume={35},
  pages={36479--36494},
  year={2022}
}

@article{dhariwal2021diffusion,
  title={Diffusion models beat gans on image synthesis},
  author={Dhariwal, Prafulla and Nichol, Alexander},
  journal={Advances in neural information processing systems},
  volume={34},
  pages={8780--8794},
  year={2021}
}

@article{gal2023encoder,
  title={Encoder-based domain tuning for fast personalization of text-to-image models},
  author={Gal, Rinon and Arar, Moab and Atzmon, Yuval and Bermano, Amit H and Chechik, Gal and Cohen-Or, Daniel},
  journal={ACM Transactions on Graphics (TOG)},
  volume={42},
  number={4},
  pages={1--13},
  year={2023},
  publisher={ACM New York, NY, USA}
}

@article{chen2023disenbooth,
  title={Disenbooth: Identity-preserving disentangled tuning for subject-driven text-to-image generation},
  author={Chen, Hong and Zhang, Yipeng and Wu, Simin and Wang, Xin and Duan, Xuguang and Zhou, Yuwei and Zhu, Wenwu},
  journal={arXiv preprint arXiv:2305.03374},
  year={2023}
}

@article{gal2022image,
  title={An image is worth one word: Personalizing text-to-image generation using textual inversion},
  author={Gal, Rinon and Alaluf, Yuval and Atzmon, Yuval and Patashnik, Or and Bermano, Amit H and Chechik, Gal and Cohen-Or, Daniel},
  journal={arXiv preprint arXiv:2208.01618},
  year={2022}
}

@article{ye2023ip,
  title={Ip-adapter: Text compatible image prompt adapter for text-to-image diffusion models},
  author={Ye, Hu and Zhang, Jun and Liu, Sibo and Han, Xiao and Yang, Wei},
  journal={arXiv preprint arXiv:2308.06721},
  year={2023}
}

@inproceedings{ronneberger2015u,
  title={U-net: Convolutional networks for biomedical image segmentation},
  author={Ronneberger, Olaf and Fischer, Philipp and Brox, Thomas},
  booktitle={Medical image computing and computer-assisted intervention--MICCAI 2015: 18th international conference, Munich, Germany, October 5-9, 2015, proceedings, part III 18},
  pages={234--241},
  year={2015},
  organization={Springer}
}

@inproceedings{chen2024training,
  title={Training-free layout control with cross-attention guidance},
  author={Chen, Minghao and Laina, Iro and Vedaldi, Andrea},
  booktitle={Proceedings of the IEEE/CVF Winter Conference on Applications of Computer Vision},
  pages={5343--5353},
  year={2024}
}

@article{li2024blip,
  title={Blip-diffusion: Pre-trained subject representation for controllable text-to-image generation and editing},
  author={Li, Dongxu and Li, Junnan and Hoi, Steven},
  journal={Advances in Neural Information Processing Systems},
  volume={36},
  year={2024}
}

@article{liu2023cones,
  title={Cones: Concept neurons in diffusion models for customized generation},
  author={Liu, Zhiheng and Feng, Ruili and Zhu, Kai and Zhang, Yifei and Zheng, Kecheng and Liu, Yu and Zhao, Deli and Zhou, Jingren and Cao, Yang},
  journal={arXiv preprint arXiv:2303.05125},
  year={2023}
}

@article{gu2024mix,
  title={Mix-of-show: Decentralized low-rank adaptation for multi-concept customization of diffusion models},
  author={Gu, Yuchao and Wang, Xintao and Wu, Jay Zhangjie and Shi, Yujun and Chen, Yunpeng and Fan, Zihan and Xiao, Wuyou and Zhao, Rui and Chang, Shuning and Wu, Weijia and others},
  journal={Advances in Neural Information Processing Systems},
  volume={36},
  year={2024}
}

@article{wang2024event,
  title={Event-customized image generation},
  author={Wang, Zhen and Jiang, Yilei and Zheng, Dong and Xiao, Jun and Chen, Long},
  journal={arXiv preprint arXiv:2410.02483},
  year={2024}
}

@inproceedings{han2023svdiff,
  title={Svdiff: Compact parameter space for diffusion fine-tuning},
  author={Han, Ligong and Li, Yinxiao and Zhang, Han and Milanfar, Peyman and Metaxas, Dimitris and Yang, Feng},
  booktitle={Proceedings of the IEEE/CVF International Conference on Computer Vision},
  pages={7323--7334},
  year={2023}
}

@inproceedings{kwon2024concept,
  title={Concept weaver: Enabling multi-concept fusion in text-to-image models},
  author={Kwon, Gihyun and Jenni, Simon and Li, Dingzeyu and Lee, Joon-Young and Ye, Jong Chul and Heilbron, Fabian Caba},
  booktitle={Proceedings of the IEEE/CVF Conference on Computer Vision and Pattern Recognition},
  pages={8880--8889},
  year={2024}
}

@inproceedings{chen2025iteris,
  title={Iteris: Iterative inference-solving alignment for lora merging},
  author={Chen, Hongxu and Wang, Zhen and Li, Runshi and Zhu, Bowei and Chen, Long},
  booktitle={Proceedings of the Computer Vision and Pattern Recognition Conference},
  pages={4829--4838},
  year={2025}
}

@article{liu2023customizable,
  title={Customizable image synthesis with multiple subjects},
  author={Liu, Zhiheng and Zhang, Yifei and Shen, Yujun and Zheng, Kecheng and Zhu, Kai and Feng, Ruili and Liu, Yu and Zhao, Deli and Zhou, Jingren and Cao, Yang},
  journal={Advances in neural information processing systems},
  volume={36},
  pages={57500--57519},
  year={2023}
}

@inproceedings{zhu2025multibooth,
  title={Multibooth: Towards generating all your concepts in an image from text},
  author={Zhu, Chenyang and Li, Kai and Ma, Yue and He, Chunming and Li, Xiu},
  booktitle={Proceedings of the AAAI Conference on Artificial Intelligence},
  volume={39},
  number={10},
  pages={10923--10931},
  year={2025}
}

@inproceedings{ding2024freecustom,
  title={Freecustom: Tuning-free customized image generation for multi-concept composition},
  author={Ding, Ganggui and Zhao, Canyu and Wang, Wen and Yang, Zhen and Liu, Zide and Chen, Hao and Shen, Chunhua},
  booktitle={Proceedings of the IEEE/CVF Conference on Computer Vision and Pattern Recognition},
  pages={9089--9098},
  year={2024}
}

@inproceedings{reed2016generative,
  title={Generative adversarial text to image synthesis},
  author={Reed, Scott and Akata, Zeynep and Yan, Xinchen and Logeswaran, Lajanugen and Schiele, Bernt and Lee, Honglak},
  booktitle={International conference on machine learning},
  pages={1060--1069},
  year={2016},
  organization={Pmlr}
}

@article{li2019controllable,
  title={Controllable text-to-image generation},
  author={Li, Bowen and Qi, Xiaojuan and Lukasiewicz, Thomas and Torr, Philip},
  journal={Advances in neural information processing systems},
  volume={32},
  year={2019}
}

@article{song2020denoising,
  title={Denoising diffusion implicit models},
  author={Song, Jiaming and Meng, Chenlin and Ermon, Stefano},
  journal={arXiv preprint arXiv:2010.02502},
  year={2020}
}

@article{ho2022classifier,
  title={Classifier-free diffusion guidance},
  author={Ho, Jonathan and Salimans, Tim},
  journal={arXiv preprint arXiv:2207.12598},
  year={2022}
}

@inproceedings{rombach2022high,
  title={High-resolution image synthesis with latent diffusion models},
  author={Rombach, Robin and Blattmann, Andreas and Lorenz, Dominik and Esser, Patrick and Ommer, Bj{\"o}rn},
  booktitle={Proceedings of the IEEE/CVF conference on computer vision and pattern recognition},
  pages={10684--10695},
  year={2022}
}

@article{podell2023sdxl,
  title={Sdxl: Improving latent diffusion models for high-resolution image synthesis},
  author={Podell, Dustin and English, Zion and Lacey, Kyle and Blattmann, Andreas and Dockhorn, Tim and M{\"u}ller, Jonas and Penna, Joe and Rombach, Robin},
  journal={arXiv preprint arXiv:2307.01952},
  year={2023}
}

@article{li2023blip,
  title={Blip-diffusion: Pre-trained subject representation for controllable text-to-image generation and editing},
  author={Li, Dongxu and Li, Junnan and Hoi, Steven},
  journal={Advances in Neural Information Processing Systems},
  volume={36},
  pages={30146--30166},
  year={2023}
}

@inproceedings{zhang2024ssr,
  title={Ssr-encoder: Encoding selective subject representation for subject-driven generation},
  author={Zhang, Yuxuan and Song, Yiren and Liu, Jiaming and Wang, Rui and Yu, Jinpeng and Tang, Hao and Li, Huaxia and Tang, Xu and Hu, Yao and Pan, Han and others},
  booktitle={Proceedings of the IEEE/CVF Conference on Computer Vision and Pattern Recognition},
  pages={8069--8078},
  year={2024}
}

@inproceedings{kong2024omg,
  title={Omg: Occlusion-friendly personalized multi-concept generation in diffusion models},
  author={Kong, Zhe and Zhang, Yong and Yang, Tianyu and Wang, Tao and Zhang, Kaihao and Wu, Bizhu and Chen, Guanying and Liu, Wei and Luo, Wenhan},
  booktitle={European Conference on Computer Vision},
  pages={253--270},
  year={2024},
  organization={Springer}
}

@inproceedings{jiang2025mc,
  title={MC\^{} 2: Multi-concept Guidance for Customized Multi-concept Generation},
  author={Jiang, Jiaxiu and Zhang, Yabo and Feng, Kailai and Wu, Xiaohe and Li, Wenbo and Pei, Renjing and Li, Fan and Zuo, Wangmeng},
  booktitle={Proceedings of the Computer Vision and Pattern Recognition Conference},
  pages={2802--2812},
  year={2025}
}

@article{gu2023mix,
  title={Mix-of-show: Decentralized low-rank adaptation for multi-concept customization of diffusion models},
  author={Gu, Yuchao and Wang, Xintao and Wu, Jay Zhangjie and Shi, Yujun and Chen, Yunpeng and Fan, Zihan and Xiao, Wuyou and Zhao, Rui and Chang, Shuning and Wu, Weijia and others},
  journal={Advances in Neural Information Processing Systems},
  volume={36},
  pages={15890--15902},
  year={2023}
}

@article{fu2023dreamsim,
  title={DreamSim: Learning New Dimensions of Human Visual Similarity using Synthetic Data},
  author={Fu, Stephanie and Tamir, Netanel and Sundaram, Shobhita and Chai, Lucy and Zhang, Richard and Dekel, Tali and Isola, Phillip},
  journal={Advances in Neural Information Processing Systems},
  volume={36},
  pages={50742--50768},
  year={2023}
}

@article{oquabdinov2,
  title={DINOv2: Learning Robust Visual Features without Supervision},
  author={Oquab, Maxime and Darcet, Timoth{\'e}e and Moutakanni, Th{\'e}o and Vo, Huy V and Szafraniec, Marc and Khalidov, Vasil and Fernandez, Pierre and HAZIZA, Daniel and Massa, Francisco and El-Nouby, Alaaeldin and others},
  journal={Transactions on Machine Learning Research}
}

@article{soboleva2025t,
  title={T-LoRA: Single Image Diffusion Model Customization Without Overfitting},
  author={Soboleva, Vera and Alanov, Aibek and Kuznetsov, Andrey and Sobolev, Konstantin},
  journal={arXiv preprint arXiv:2507.05964},
  year={2025}
}

@article{kwon2024tweediemix,
  title={Tweediemix: Improving multi-concept fusion for diffusion-based image/video generation},
  author={Kwon, Gihyun and Ye, Jong Chul},
  journal={arXiv preprint arXiv:2410.05591},
  year={2024}
}
}
\newpage


\appendix
\section*{Appendix}

This appendix is organized as follows:
\begin{itemize}
\item Section~\ref{sec:appendix-visual-attention map} provides the visualizations of attention maps and detailed analysis.

\item Section~\ref{sec:appendix-user-study} provides the user study of baselines and our method.

\item Section~\ref{sec:appendix-visualization} provides implementation details of teaser and more visualization results.

\item Section~\ref{sec:appendix-exps} provides implementation details of quantitative evaluation.
\item Section~\ref{sec:appendix-impact} provides the broader impact of our method.

\end{itemize}

\begin{figure*}
    \centering
    \includegraphics[width=1\linewidth]{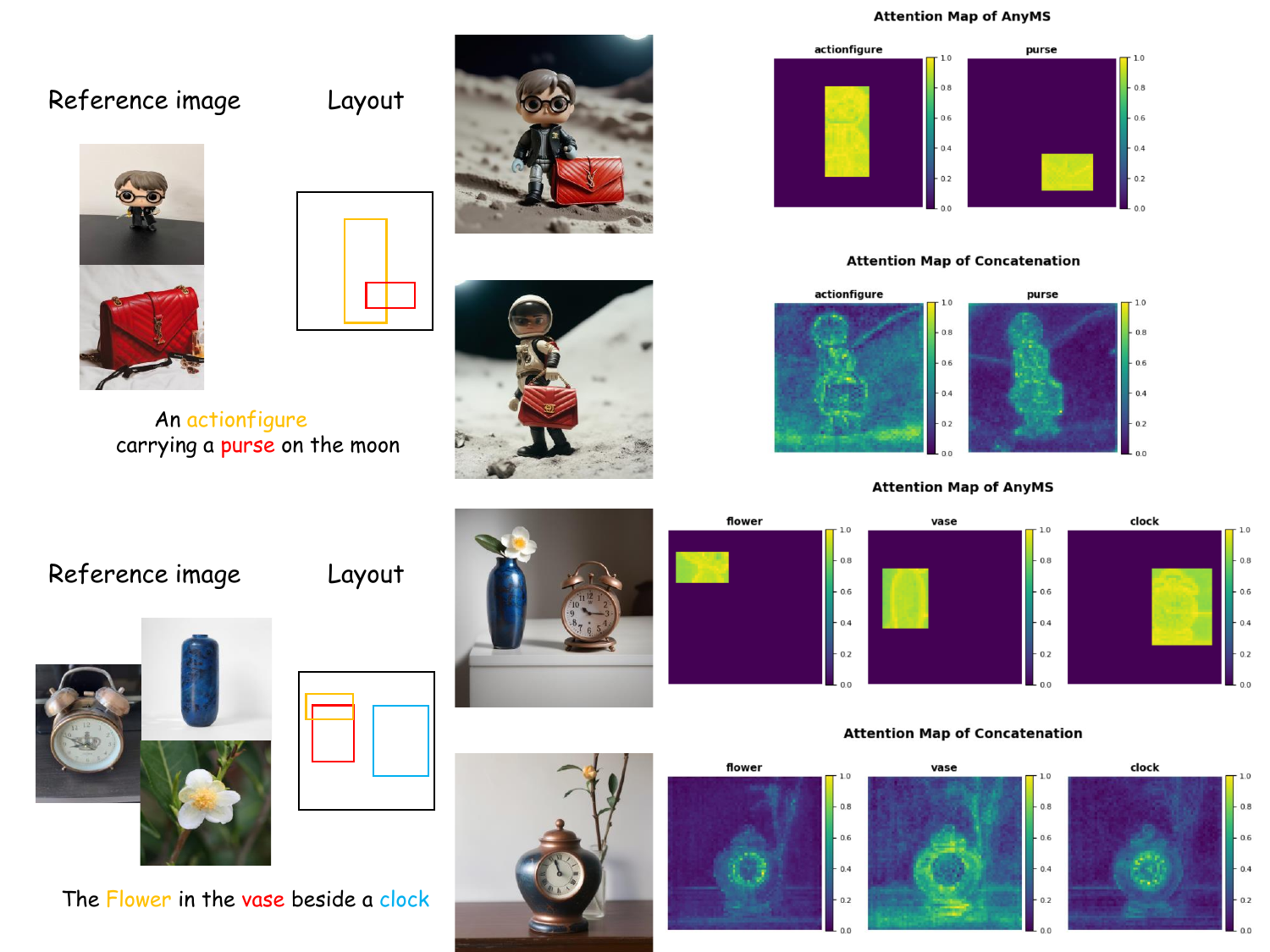}
    \caption{\textbf{Visualization of attention maps}. We compute the mean of the attention maps associated with the subjects and convert the result into normalized heatmaps.}
    \label{fig:visualization of attention map}
    \vspace{-1em}
\end{figure*}

\section{Visualization of Attention Maps and Detailed Analysis}
\label{sec:appendix-visual-attention map}
In this section, we provide visualizations of attention maps in Figure~\ref{fig:visualization of attention map} to further demonstrate the effectiveness of our method. As stated in the main text, directly concatenating image and text features and processing them through text cross-attention leads to entanglement among features of different subjects evidenced by overlapping highlighted regions in attention maps and then results in issues such as object omission, fidelity degradation, and unintended attribute transfer. AnyMS addresses this through dual-level attention decoupling, ensuring that each subject attends only to its corresponding spatial region. This simultaneously disentangles features of different subjects  and enables precise layout control. Furthermore, global decoupling is employed to maintain semantic consistency across the generation.

\begin{figure*}[!t]
    \centering
    \includegraphics[width=1\linewidth]{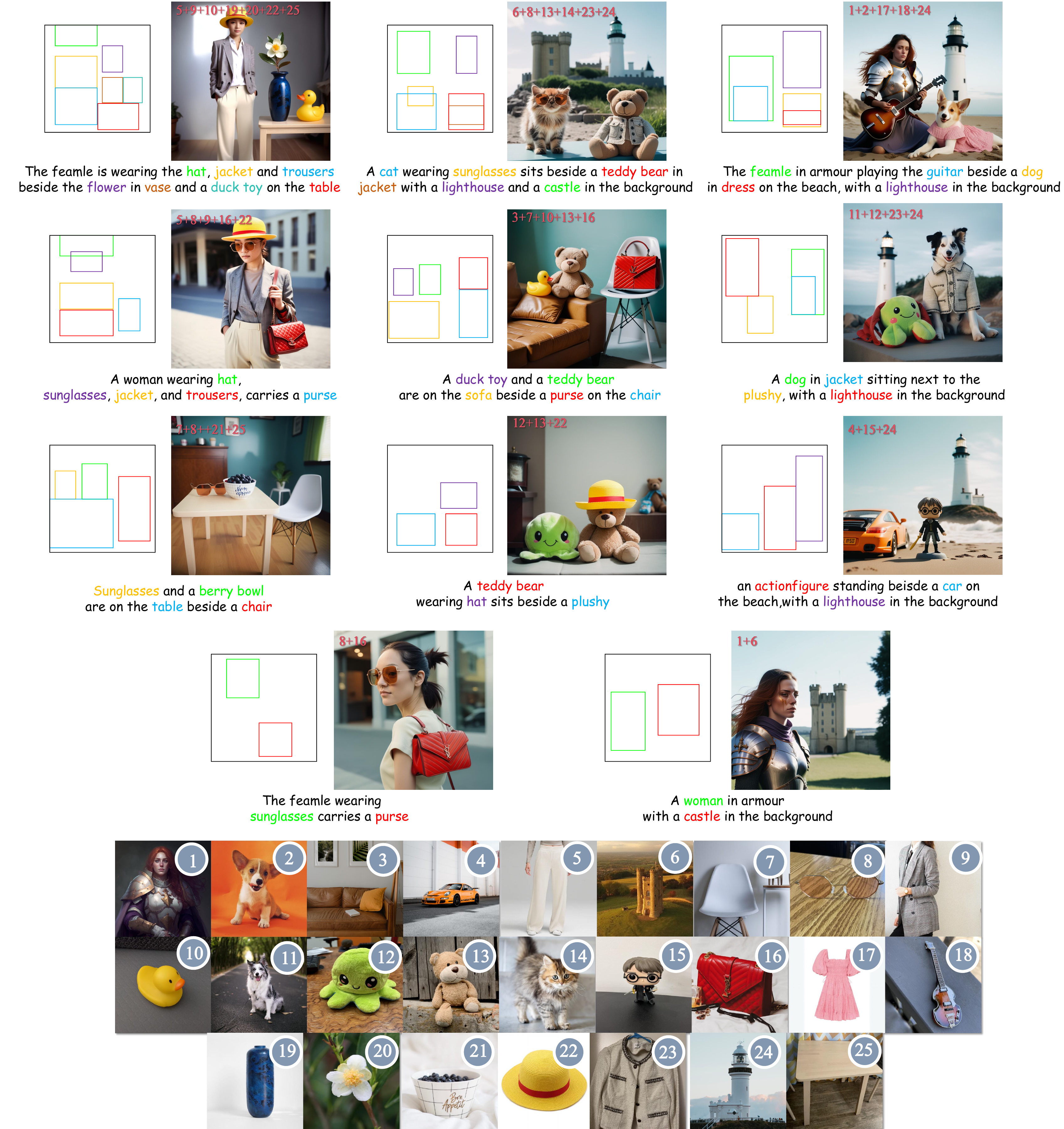}
    \vspace{-2em}
    \caption{\textbf{Visualization Details and Additional Results.} Different colors show the associations between subjects and their corresponding layout configurations. We show more results beyond the first page with these settings.}
    \label{fig:details of first}
    \vspace{-1em}
\end{figure*}

\section{User Study}
\label{sec:appendix-user-study}
We further evaluated the proposed AnyMS method through a user study. The study consisted of 9 sets of generation cases, with 25 participants in total. In each case, users were presented with a text prompt, reference images, layout requirements, and corresponding results generated by different methods. Participants are asked to select one or two images they consider best aligned with these four criteria. respectively : 1) presence of the target concepts and their alignment with the reference images; 2) alignment with the given text prompt; 3) adherence to the specified layout; 4) overall quality and authenticity of the generated images. We then use the proportion of votes each baseline received out of the total number of votes as the score to measure the performance of each method. The results are shown in Table~\ref{tab:user_study}. Our method attained the highest scores across all metrics and overall visual impression, demonstrating a strong user preference, and achieves a well-balanced trade-off among layout control, text alignment, and identity preservation.
\begin{table}[H]
\centering
\small
\begin{tabular}{|c|c|c|c|c|}
\hline
Model & \makecell{Concept \\ Alignment}  & \makecell{Text \\ Alignment}  & \makecell{Layout \\ Control}  & \makecell{Overall \\ Quality} \\
\hline
Cones 2 & 0.08 & 0.07 & 0.05 & 0.05 \\
MuDI & 0.17 & 0.06 & 0.14 & 0.05 \\
MS-Diffusion & 0.18 & 0.22 & 0.23 & 0.18 \\
AnyMS(Ours) & \textbf{0.57}& \textbf{0.65} & \textbf{0.58} & \textbf{0.72} \\
\hline
\end{tabular}
\caption{\textbf{User Study.} Scored by the fraction of total votes received by each baseline. Our method obtains the best results on all four dimensions, demonstrating superior overall quality.}
\label{tab:user_study}
\end{table}

\section{Visualization Details and More Results}
\label{sec:appendix-visualization}
We provide implementation details of teaser on the first page, including layout configurations and text prompts, and we also show additional multi-subject customization results based on the settings in Figure~\ref{fig:details of first}. The results encompass various combination types and include scenarios with the number of subjects ranging from 2 to 7, fully demonstrating the generalizability and robustness of AnyMS. AnyMS has the ability to balance the trade-off among text-alignment, subject identity preservation, and layout control while scaling to a larger number of subjects.

\section{Implementation Details of Quantitative Evaluation}
\label{sec:appendix-exps}

\begin{figure*}[t]
    \centering
    \includegraphics[width=1\linewidth]{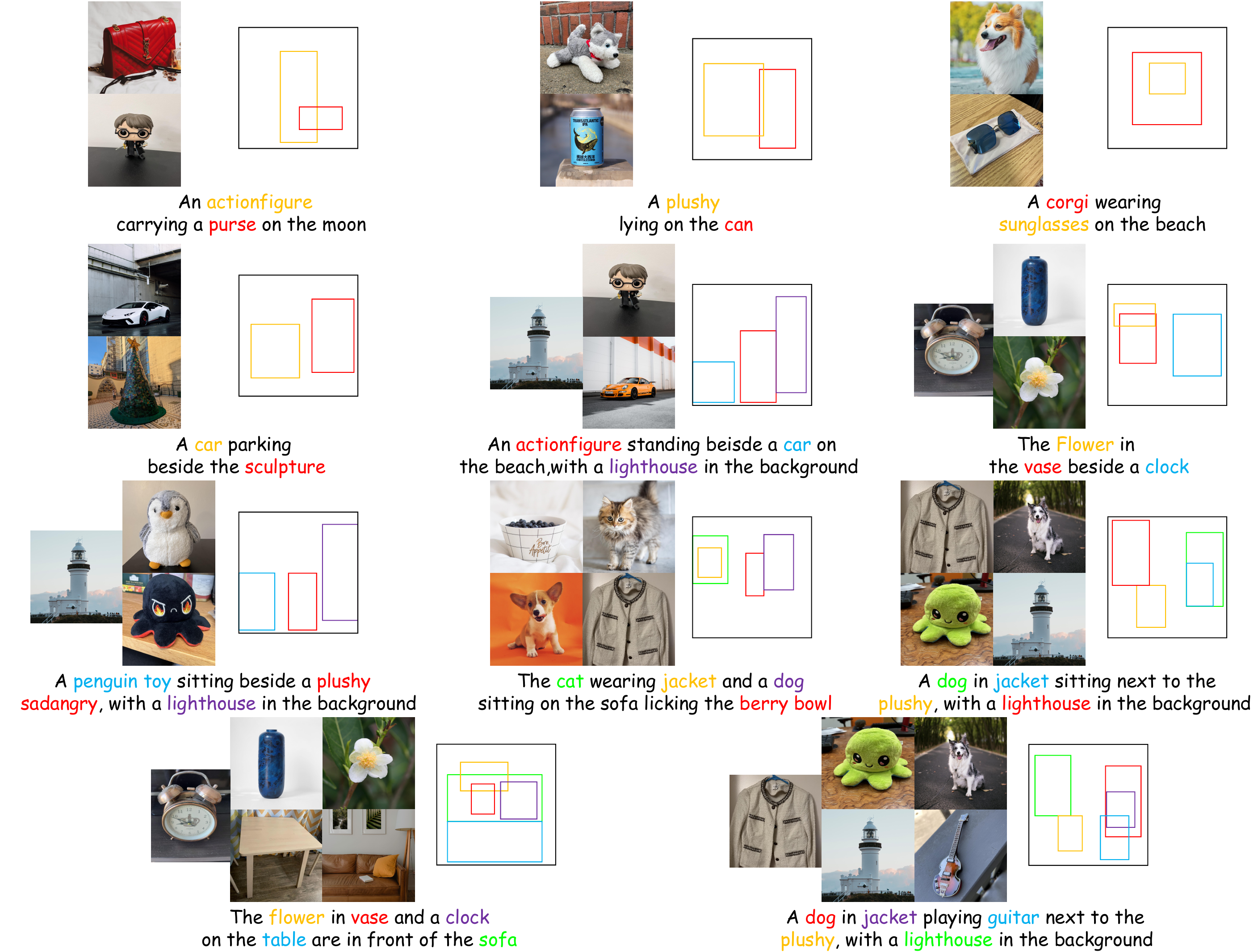}
    \caption{\textbf{Details of 11 Combinations in Quantitative Evaluation.} Different colors show the associations between subjects and their corresponding layout configurations.}
    \label{fig:details of 11 combination}
    \vspace{-1em}
\end{figure*}

As shown in Figure~\ref{fig:details of 11 combination}, for the quantitative evaluation, we collect 24 different subjects covering a variety of categories, and form 11 combinations. For each combination, we specify a tuple containing reference images, text-prompt, and bounding boxes. To conduct a comprehensive evaluation, the number of subjects ranged from 2 to 5, assessing the ability of models to decouple subject features as the number of subjects increases. In addition, the ways subjects interact (e.g., carrying and wearing) and the backgrounds (e.g., moon and beach) are also diverse.

\section{Broader Impact}
\label{sec:appendix-impact}
As a multi-subject customization method, AnyMS can generate a text-aligned and well-composed image involving multiple user-provided subjects without training. This means that AnyMS has the potential to play a crucial role in advertising and the film industries: we can create digital doubles of actors and seamlessly render them with virtual cartoon characters into posters for propagation. In addition, generated images with a variety of backgrounds and styles can further enhance diversity, offering audiences a novel experience. AnyMS can also function as entertainment for users in social media. However, there are several points to note when implementing AnyMS. 1) Sensitive terms such as sexual content and political statements should be detected and blocked. 2) Privacy and portrait rights of individuals should be protected. The generated images should be authorized by the person concerned. Overall, AnyMS can enrich entertainment activities for the public with proper application.

\end{document}